\documentclass[conference, final]{IEEEtran}
\IEEEoverridecommandlockouts
\usepackage{cite}
\usepackage{amsmath,amssymb,amsfonts}
\usepackage{algorithmic}
\usepackage{graphicx}
\usepackage{textcomp}
\usepackage{xcolor}
\usepackage{times}
\usepackage{soul}
\usepackage{url}
\usepackage[hidelinks]{hyperref}
\usepackage[utf8]{inputenc}
\usepackage[small]{caption}
\usepackage{graphicx}
\usepackage{amsmath}
\usepackage{amsthm}
\usepackage{array}
\usepackage{booktabs}
\usepackage{algorithm}
\usepackage{algorithmic}
\usepackage[switch]{lineno}

\usepackage{algorithm}
\usepackage{algorithmic}
\def\method{\textsc{DGOTTA}}
\usepackage{amsmath}
\usepackage{booktabs}
\usepackage{caption}
\usepackage{subcaption}
\usepackage{amsfonts}
\usepackage{color}
\usepackage{tikz}
\usepackage{enumitem}
\usepackage{pifont}
\newcolumntype{P}[1]{>{\centering\arraybackslash}p{#1}}
\usepackage[utf8]{inputenc}

\def\BibTeX{{\rm B\kern-.05em{\sc i\kern-.025em b}\kern-.08em
    T\kern-.1667em\lower.7ex\hbox{E}\kern-.125emX}}
\begin{document}

\title{Temporal Memory-Aware Online Test-Time Adaptation on Dynamic Graphs}
\IEEEoverridecommandlockouts

\author{
\IEEEauthorblockN{
Bo Li\textsuperscript{1},
Xin Zheng\textsuperscript{2},
Ming Jin\textsuperscript{1},
Can Wang\textsuperscript{1},
Shirui Pan\textsuperscript{1,*}\thanks{* Corresponding author.}
}
\IEEEauthorblockA{
\textsuperscript{1} Griffith University, Gold Coast, Australia\\
\textsuperscript{2}RMIT University, Melbourne, Australia\\
\{bo.li, ming.jin, can.wang, s.pan\}@griffith.edu.au,
xin.zheng2@rmit.edu.au
}
}


\maketitle

\begin{abstract}
Test-time adaptation (TTA) on graphs aims to adapt a graph neural network (GNN) that is well-trained on the training graph to the test graph, which involves potential distribution shifts that may harm model generalization and test-time inference.
While recent efforts have investigated TTA on static graphs, there is still a research gap on dynamic graphs learned with dynamic GNN (DGNN) models, where both structural connectivity and node semantics evolve continuously over time. This makes adapting a DGNN model for reliable test-time performance substantially challenging.
To fill this gap, in this work, we propose a novel framework of temporal memory-aware \underline{\textbf{O}}nline \underline{\textbf{T}}est-\underline{\textbf{T}}ime \underline{\textbf{A}}daptation on \underline{\textbf{D}}ynamic \underline{\textbf{G}}raphs, named~\textbf{\method}, to effectively adapt well-trained DGNNs during test time.
Specifically, the proposed~\method~contains three modules: (1) \textit{temporal-aware augmentation}, to extend the diversity of test dynamic graphs for addressing complex temporal and spatial shifts; (2) \textit{memory-aware model prediction}, to alleviate catastrophic forgetting; (3) \textit{consistency-guided online adaptation}, to enforce temporal alignment and memory smoothness. Extensive experiments on three real-world datasets and four DGNN backbones demonstrate that \method~significantly improves generalization under diverse distribution shifts and multiple model architectures.
\end{abstract}
\section{Introduction}
Dynamic graphs are prevalent in real-world applications such as social networks~\cite{berger2006framework,greene2010tracking,sun2022aligning,song2019session}, financial transaction networks~\cite{nascimento2021dynamic,zhang2021dyngraphtrans}, and traffic systems~\cite{li2023dynamic,zhou2020foresee,zhou2020riskoracle,lan2022dstagnn}, where both the topology and node attributes evolve over time. Dynamic graph neural networks (DGNN), have been recognized as powerful models for capturing the complex interplay of spatio-temporal dependencies, and learning expressive dynamic node and interaction representations.

In this work, we characterize distribution shift in dynamic graphs as the discrepancy between training-time and test-time graph streams along both temporal and structural dimensions. Concretely, such shifts manifest as changes in graph density, node feature statistics, and edge dynamics as the graph evolves over time, rather than as i.i.d. feature perturbations.

\begin{figure*}[t]
\centering

\includegraphics[width=0.95\linewidth]{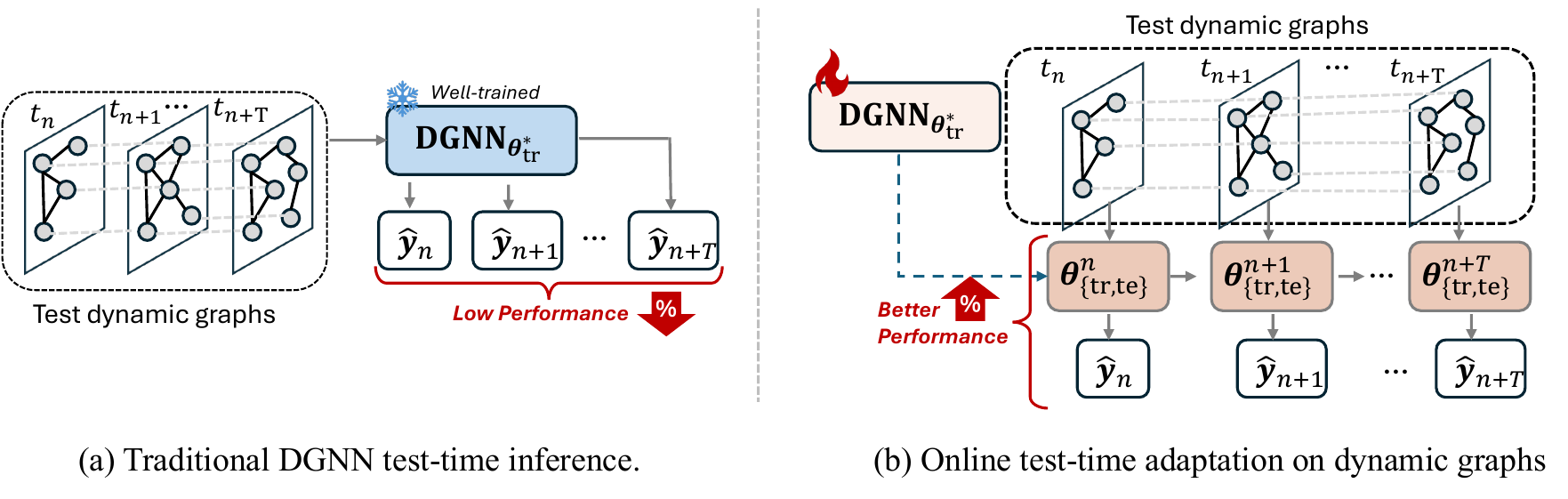}


\begin{tabular}{ccc}
    \includegraphics[width=0.30\linewidth]{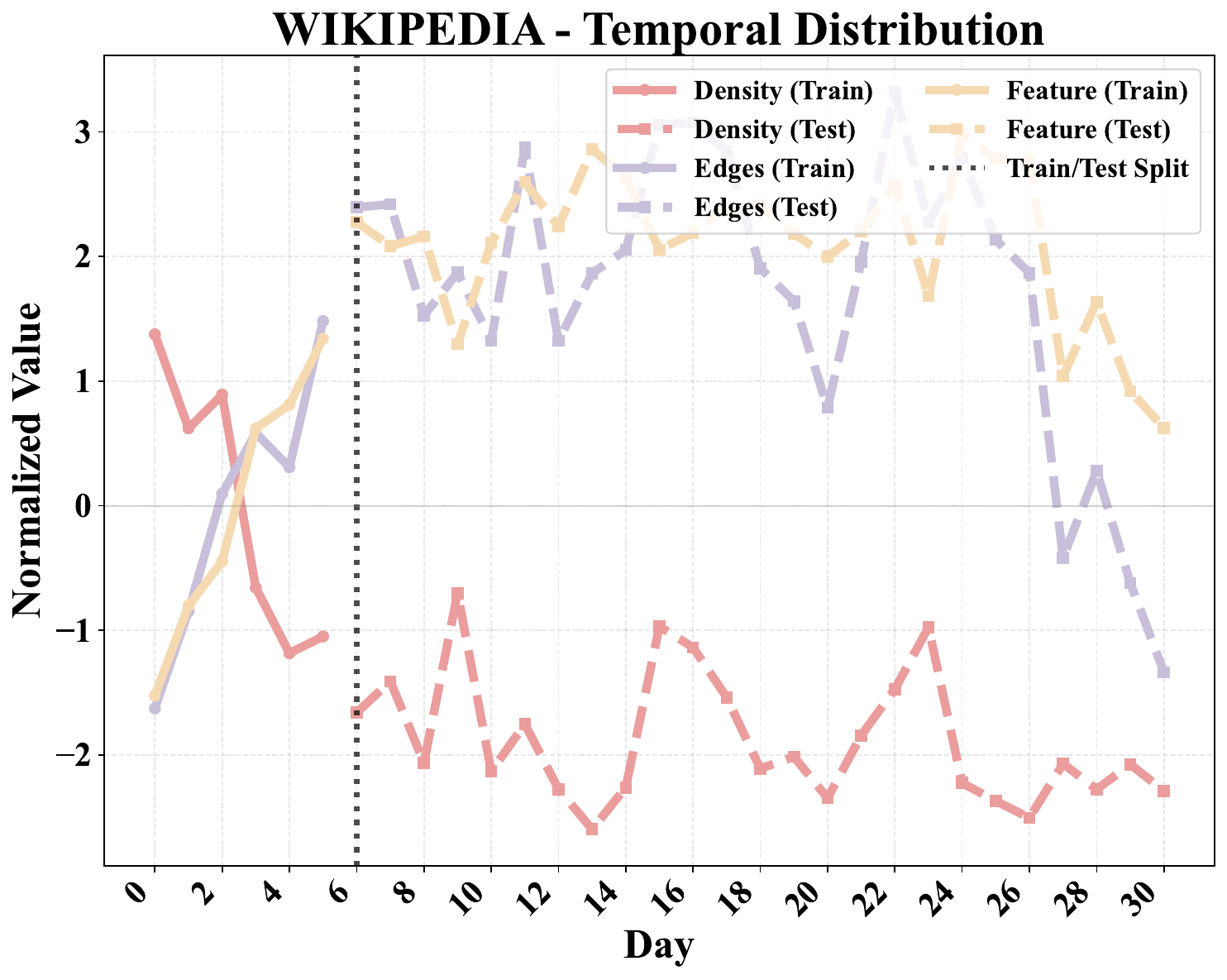} &
    \includegraphics[width=0.30\linewidth]{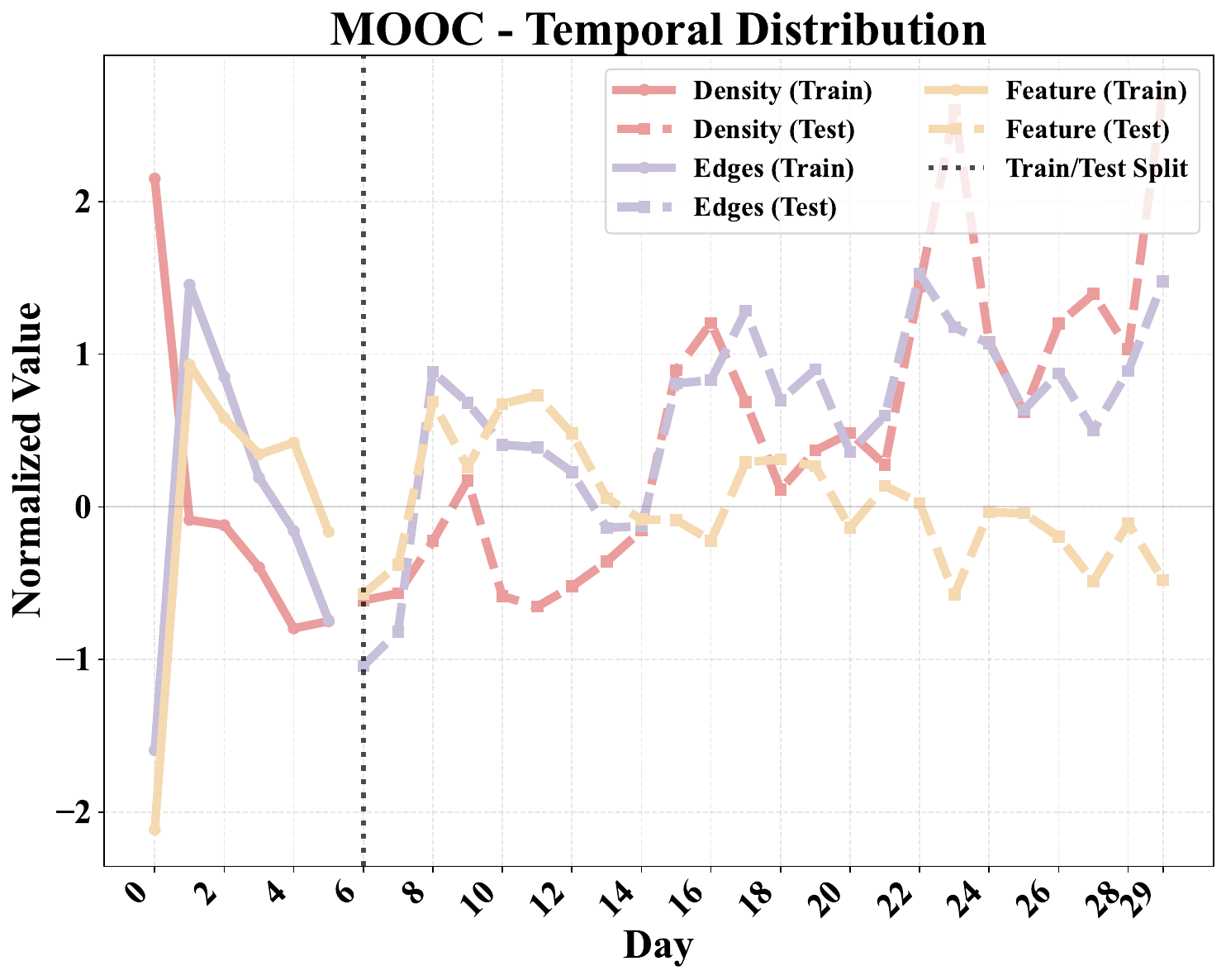} &
    \includegraphics[width=0.30\linewidth]{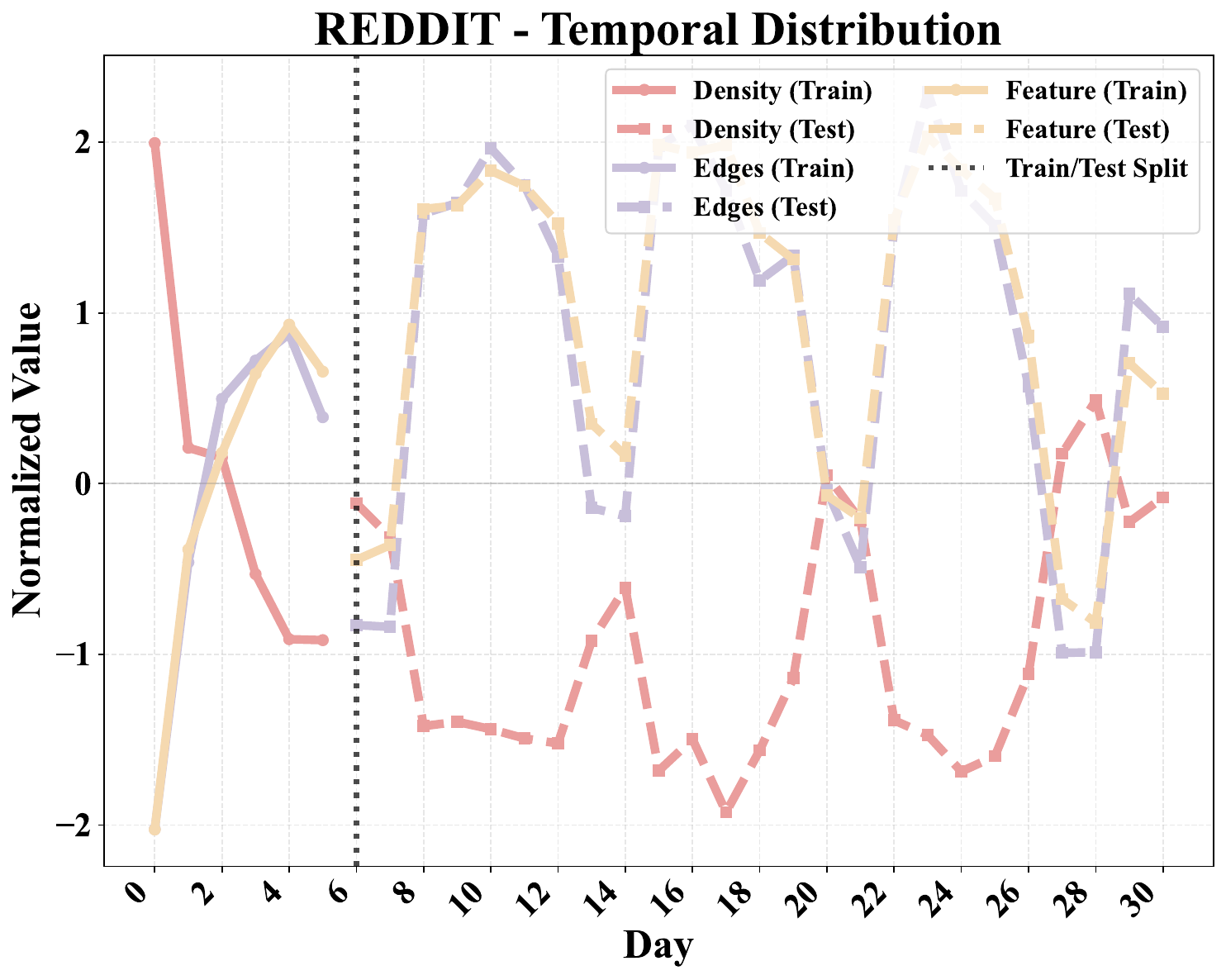}
\end{tabular}
\caption{Motivation and empirical evidence of temporal distribution shift in dynamic graphs.
(a) Traditional DGNN test-time inference, where a well-trained model with fixed parameters is directly deployed on evolving test graph streams, leading to degraded performance under temporal and structural shifts.
(b) Online test-time adaptation on dynamic graphs, which continuously updates model parameters during deployment to better align with evolving test distributions. Empirical visualization of temporal distribution discrepancies between training and test graph snapshots on Wikipedia, MOOC, and Reddit, respectively.}
\label{fig:temporal_shift}
\end{figure*}

As illustrated in Figure~\ref{fig:temporal_shift}, real-world dynamic graph benchmarks such as Wikipedia, MOOC, and Reddit exhibit clear temporal discrepancies between training and test snapshots across these statistics. The test-time graph streams demonstrate evolving patterns in interaction frequency, feature distributions, and connectivity dynamics, indicating that the underlying data distribution is non-stationary and continuously drifting. These observations highlight that distribution shift in dynamic graphs is inherently temporal and structural, which fundamentally challenges the assumption of stationary test distributions adopted by conventional DGNN inference.

Moreover, DGNNs that are well-trained on the training dynamic graphs usually show poor performance and generalization ability to new incoming unlabeled test graphs~\cite{yuan2025dg}, due to potential distribution shifts that might harm the performance. For instance, in recommendation systems where dynamic graphs capture user-item interactions, changes in user features (e.g., an increase income) can significantly alter user preferences and behaviors (e.g., a shift towards higher-priced items). Consequently, a DGNN model trained on earlier feature distributions (e.g., lower income) may struggle to deliver relevant recommendations at test time, as it relies on outdated interactions (such as purchases of lower-cost items)~\cite{yuan2024environment,wang2022causal}.
Despite the effectiveness of well-trained DGNNs on the training graphs, their generalization capability often deteriorates on distribution-shifted test graphs, especially when deployed in realistic dynamic environments.

As shown in Figure~\ref{fig:temporal_shift}, a well-trained DGNN with fixed parameters $\boldsymbol{\theta}^*_{\mathrm{tr}}$ is directly applied to the test phase, where test dynamic graphs are inherently evolving—node semantics shift, new edges appear, and structural patterns change continuously. As a result, the traditional DGNN inference would lead to low performance with poor generalization.

To address this test-time degradation problem, test-time adaptation (TTA) 
methods on graphs, which aim to adapt a well-trained GNN to unseen test distributions, become a promising solution. 
However, existing TTA methods on graphs mainly focus on static graphs. There is still a research blank for exploring TTA on dynamic graphs with DGNNs. Specifically, there are two core challenges for TTA with DGNNs: \textbf{Challenge-1}: at the dynamic graph level, complex temporal and spatial shifts on test graphs complicate the informative representation transfer during test time, leading to poor generalization; \textbf{Challenge-2}: at the method level, temporal evolution enforces an online learning paradigm in DGNNs, where catastrophic forgetting severely degrades DGNNs' test-time inference performance.


%

To address these challenges, in this work, we propose a novel temporal memory-aware \underline{\textbf{\textsc{O}}}nline \underline{\textbf{\textsc{T}}}est-\underline{\textbf{\textsc{T}}}ime \underline{\textbf{\textsc{A}}}daptation method for \underline{\textbf{\textsc{D}}}ynamic \underline{\textbf{\textsc{G}}}raphs, named~\textbf{\method}, to effectively adapt well-trained DGNNs to evolving test-time dynamic graphs with potential distribution shifts. 
Specifically, the proposed method contains three essential modules:
(1) \textit{temporal-aware augmentation}, to extend the diversity of test dynamic graphs for addressing complex temporal and spatial shifts, leading to improved robustness against temporal shifts (C1); (2) \textit{memory-aware model prediction}, to stabilize unsupervised adaptation with the exponential moving average strategy and alleviate catastrophic forgetting (C2); and (3) \textit{consistency-guided online adaptation}, to enforce prediction alignment across time and ensure smooth adaptation.
By doing so, the proposed method could enable reliable online adaptation to dynamic graphs without access to labeled data. Extensive experiments on three real-world datasets and four DGNN backbones demonstrate that \method~significantly improves generalization under diverse distribution shifts and multiple model architectures. In summary, the contributions of this work are presented as:
\begin{itemize}
    \item To the best of our knowledge, we are the first to study online test-time adaptation for dynamic graphs, enabling models to adapt temporal and structural distribution shifts during inference in a model-agnostic manner.
    \item We develop a novel framework named~\method, containing (1) temporal-aware augmentation; (2) memory-aware model prediction; and (3) consistency-guided online adaptation for addressing key challenges in test-time adaptation on dynamic graphs with DGNNs.
    \item Extensive experiments on real-world datasets and dynamic GNN backbones show that \method~consistently outperforms existing baselines, achieving state-of-the-art performance under dynamic distribution shifts.
\end{itemize}
\section{Related Works}

\noindent\textbf{Online Test-Time Adaptation.} OTTA aims to adapt a pretrained model from the source domain to an unlabeled target domain without re-accessing the source domain during adaptation~\cite{liang2025comprehensive}, and it allows the model to adapt the test data in a source-free and online manner~\cite{jain2011online,sun2020test,wang2020tent}. For image data, recent works have proposed image TTA by entropy minimization~\cite{wang2020tent,zhang2022memo,niu2023towards,jang2022test}, pseudo-labeling~\cite{zhang2022memo,marsden2024universal,jang2022test,zeng2024rethinking} and other like Laplacian Adjusted Maximum likelihood Estimation~\cite{boudiaf2022parameter}. A subset of OTTA approaches has explored continual or online adaptation settings. Studies on this topic focus on the problems which target domains are not fixed but change continuously in an online manner~\cite{wang2022continual,gan2023decorate,niu2023towards,song2023ecotta,lee2024becotta}. 

\noindent\textbf{Graph Test-Time Adaptation.} Studies on graph test-time adaptation (GTTA) are focusing on the problems due to the co-existence of attribute shifts and structure shifts~\cite{chen2026test, zheng2025test, zheng2025testtime}. To cope with these issues, Gtrans~\cite{jin2022empowering} proposes to refine the target graph at test time by minimizing a surrogate loss. SOGA~\cite{mao2024source} maximizes the mutual information between model inputs and outputs, and encourages consistency between neighboring or structurally similar nodes, but it is only applicable to homophilic graphs~\cite{zheng2026graph}. In terms of degree shift, GraphPatcher~\cite{ju2023graphpatcher} earns to generate virtual nodes to improve the prediction on low-degree nodes. Moreover, GAPGC~\cite{chen2022graphtta} and GT3~\cite{wang2022powerful} follow a self-supervised learning (SSL) scheme to fine-tune the pre-trained model for graph classification. There are also increasing studies on continual learning \cite{wang2024TPAMI}. The online TTA studied in this paper differs from continual learning in that the latter updates model parameters through the fine-tuning of the model with ground-truth labels, whereas online TTA has no access to labels during test time.

\noindent\textbf{Dynamic Graph Neural Networks.}
Dynamic graph neural networks aim to model graphs whose structures and attributes evolve over time. Existing DGNN methods usually encode temporal interactions through recurrent memory modules, temporal attention, time encoding, or message-passing mechanisms over continuous-time event streams~\cite{feng2025comprehensive, zheng2025survey, jin2022neural}. Representative models such as TGAT~\cite{xu2020inductive}, TGN~\cite{rossi2020temporal}, GraphMixer~\cite{cong2023we}, and DyGFormer~\cite{yu2023towards} have shown strong ability in learning temporal node representations and predicting future interactions or node labels. 
However, most DGNNs are trained under the assumption that the training and test graph streams follow similar temporal patterns. When the test stream exhibits distribution shifts, such as changes in interaction frequency, structural density, or feature statistics, a fixed DGNN may produce outdated representations and unreliable predictions~\cite{li2025test, zheng2023gnnevaluator}. Different from conventional DGNN training, our work focuses on adapting a pretrained DGNN during deployment using only unlabeled test-time graph streams. This setting is more challenging because the model must update online without supervision while preserving temporal consistency.
\section{Problem Definition}
We consider the problem of online test-time adaptation for dynamic graphs.
Given a dynamic graph stream $\{ \mathcal{G}_t = (\mathcal{V}_t, \mathcal{E}_t, \mathbf{X}_t) \}_{t=1}^T$, where $\mathcal{V}_t$ denotes the set of nodes at time $t$, $\mathcal{E}_t \subseteq \mathcal{V}_t \times \mathcal{V}_t$ is the set of timestamped edges, and $\mathbf{X}_t \in \mathbb{R}^{|\mathcal{V}_t| \times d}$ represents the node features. The problem of online test-time adaptation on dynamic graphs aims to make reliable predictions on each unseen test graph snapshot $\mathcal{G}_t$ by adjusting well-trained DGNN models.

Formally, let $f_{\boldsymbol{\theta}}$ be a well-trained DGNN model. At each test time step $t$, the model receives an unlabeled graph snapshot $\mathcal{G}_t$ and updates its parameters $\boldsymbol{\theta}_t$ via an online adaptation strategy. The objective is to minimize the expected adaptation loss over time:
\begin{equation}
    \min_{\boldsymbol{\theta}_t} \; \mathbb{E}_{\mathcal{G}_t \sim \mathcal{D}_{\text{test}}} \left[ \mathcal{L}_{\text{OTTA}}, \left(f_{\boldsymbol{\theta}_t}(\mathcal{G}_t) \right) \right]
\end{equation}
where $\mathcal{L}_{\text{OTTA}}$ denotes the adaptation objective, such as consistency loss or temporal smoothness loss, and the model is adapted incrementally in an online fashion.
\section{Methodology}
\begin{figure*}[!t]
    \centering
    \includegraphics[width=1\textwidth]{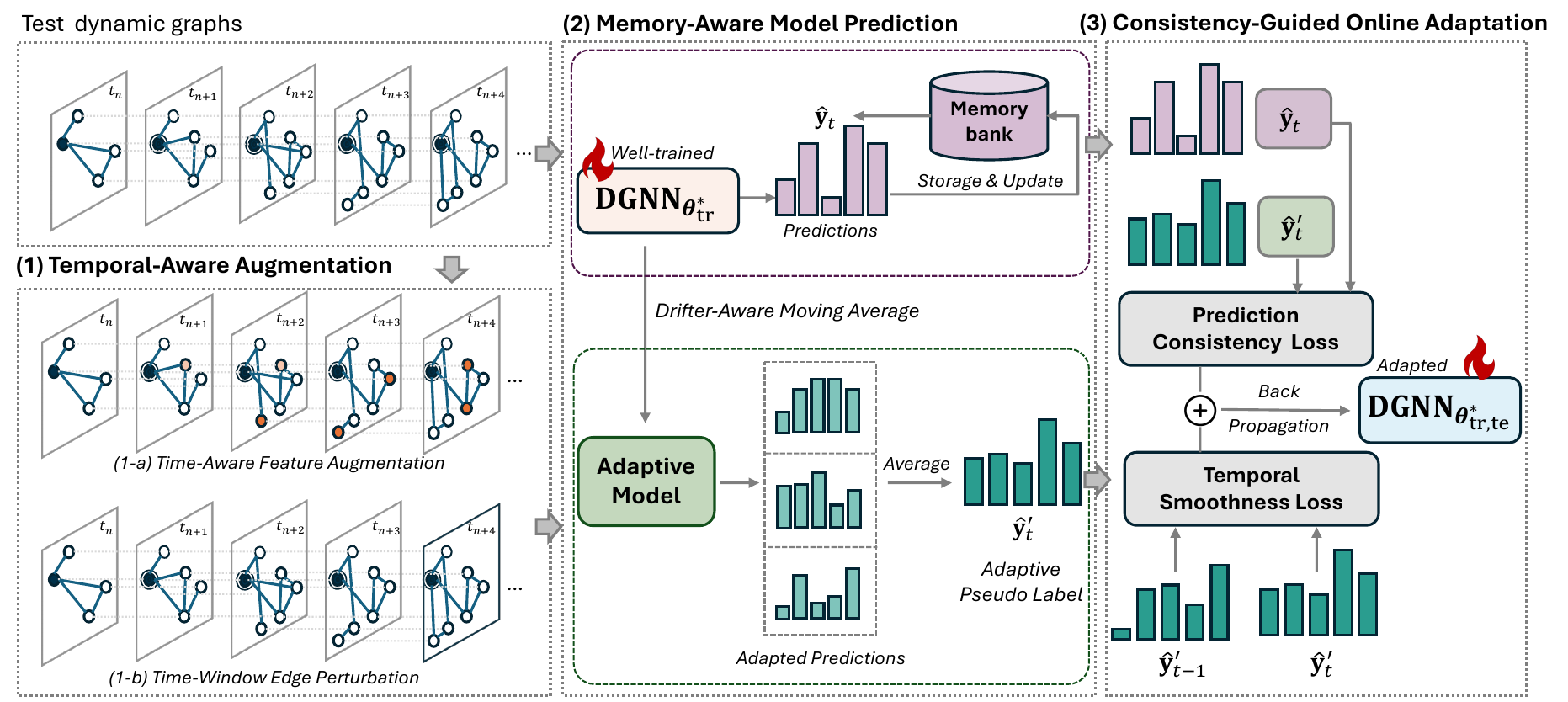}
\caption{
\textbf{Overview of our proposed temporal memory-aware online test-time adaptation for dynamic graphs.}
Given test dynamic graphs, we first conduct (1) \textit{temporal-aware augmentation} for applying time-aware feature perturbation and time-window-based edge perturbation to simulate temporal and structural drift; Then, implement (2) \textit{memory-aware model prediction} for generating soft pseudo labels, while the adaptive DGNN updates its parameters using a memory bank of past predictions; Last, develop (3) \textit{consistency-guided online adaptation} to incorporate prediction consistency and temporal smoothness losses to ensure accurate online adaptation. 
}
\label{fig:enter-label}
\end{figure*}

In this section, we introduce \method~with temporal memory-aware online test-time adaptation on dynamic graphs. The overall framework of our proposed \method~is presented in Figure~\ref{fig:enter-label}.

The design of \method follows two principles. First, the adaptation process should be temporally aware, since the usefulness of node features and edges depends strongly on their timestamps in dynamic graphs. 
This motivates temporal-aware augmentation, which produces realistic perturbed views without breaking temporal causality. Second, the adaptation process should be stable under noisy pseudo-supervision. 
Since no labels are available at test time, directly optimizing the model using its current predictions may amplify errors and cause confirmation bias. To address this issue, \method uses memory-aware pseudo labels and drift-aware parameter smoothing to balance short-term adaptability and long-term stability.

\method~enables a deployed model to adapt to temporal and structural distribution shifts by incorporating the following three key modules:

\noindent\textbf{(1) Temporal-Aware Augmentation}. To simulate distributional shifts, we introduce temporal-aware perturbations on node features and time-window-based sampling on edges, modeling both semantic drift and structural evolution.

\noindent\textbf{(2) Memory-Aware Model Prediction.} A well-trained DGNN is used to generate soft pseudo labels, while an online adaptive model continuously updates its parameters using prediction feedback. A memory bank stores recent predictions to stabilize the pseudo-label generation and smooth temporal dynamics.

\noindent\textbf{(3) Consistency-Guided Online Adaptation:} We apply prediction consistency loss between the adaptive model and pseudo labels, and a temporal consistency loss across recent predictions, jointly guiding robust online updates under evolving graph distributions.

Together, these components enable \method~to continuously self-adjust to unseen graph distributions at test time, leading to better inference performance.

\subsection{Temporal-Aware Augmentation}
\label{subsec:tta}

Graph augmentation is a crucial component of our framework, as it enhances model robustness to distribution shifts during online test-time adaptation. Unlike static graphs, temporal graphs introduce unique challenges: node interactions evolve over time and neighborhood semantics are inherently time-sensitive. Most existing augmentation strategies, however, are based on static subgraph sampling~\cite{ding2022data}, implicitly assuming that all historical interactions are equally informative. This assumption neglects the central role of time in temporal graph modeling, where the recency and order of interactions fundamentally influence message passing and prediction performance, thereby motivating the need for time-aware augmentation.

For instance, a neighbor node that interacted with the target node recently may carry much more relevant information than a node that last interacted months ago.
Naively removing such time-relevant edges or perturbing node features without considering their timestamps can distort local temporal dynamics and mislead the model during adaptation.

To address these limitations, we introduce temporal-aware augmentation, which incorporates temporal priors into both node-level and edge-level perturbations. Our augmentation simulates realistic, temporally grounded variations that help the model adapt to distribution shifts without violating the causal and temporal structure of the graph.

\noindent\textbf{Time-Aware Feature Augmentation.} In dynamic graphs, node features $\mathbf{x}_i$ may gradually become outdated or stale depending on the time elapsed since their last interaction.
To account for this temporal degradation, we design a temporal-aware augmentation mechanism that explicitly discounts older features and injects mild noise to promote robustness.

Given a node $v_i$ with raw feature vector $\mathbf{x}_i \in \mathbb{R}^d$ and last interaction timestamp $t_i$, we define the augmented feature $\widetilde{\mathbf{x}_i}$ at current time $T$ as:
\begin{equation}
\widetilde{\mathbf{x}_i} = \mathbf{x}_i \cdot e^{-\lambda(T - t_i)} + \epsilon_i,\quad \epsilon_i \sim \mathcal{N}(0, \sigma^2),
\label{eq:timeaware_feat}
\end{equation}
where $\lambda > 0$ controls the temporal decay rate and $\sigma^2$ determines the level of injected Gaussian noise.

\noindent\textbf{Time-Window Edge Perturbation.} In temporal graphs, edge relevance naturally decays over time as interactions become outdated. To reflect this temporal locality, we introduce time-window edge perturbation, which selects and perturbs edges based on their timestamps relative to the current time. Let $\mathcal{E}_t$ denote the set of edges observed up to time $t$. We first extract a subset of temporally recent edges by applying a sliding time window of size $W$:
\[
\mathcal{E}_{\text{window}} = \left\{ e_{ij} \in \mathcal{E}_t \,\middle|\, t - t_{ij} \le W \right\},
\]
where $t_{ij}$ is the timestamp of edge $e_{ij}$. To inject structural noise and simulate uncertainty in interaction dynamics, we then perform Bernoulli sampling over the retained edges with drop probability $p_{\text{drop}}$:
\begin{equation}
\widetilde{\mathcal{E}} = \mathcal{E}_{\text{window}} \,\big|\, \{ e_{ij} \sim \text{Bern}(p_{\text{drop}}) \},
\label{eq:timewindow_edge}
\end{equation}
where each edge is independently retained with probability $1 - p_{\text{drop}}$. This augmentation mechanism prevents the model from overfitting to precise temporal topologies and encourages reliance on stable, recent structures. Compared to random edge removal in static graphs, our approach is temporally informed and causality-aware, ensuring that irrelevant or outdated interactions are not mistakenly emphasized.
It is worth noting that the proposed augmentation differs from conventional random perturbation. 
In static graph augmentation, randomly dropping edges or masking features is often sufficient to improve robustness. However, in dynamic graphs, an edge is not only a structural connection but also a temporal event. 
Dropping recent edges and retaining outdated edges may distort the temporal dependency used by DGNNs for message passing. Therefore, our augmentation explicitly preserves temporal locality by first selecting recent interactions through a time window and then applying stochastic perturbation within this causally valid region. 
This design encourages the adaptive model to learn robust representations from plausible temporal variations rather than arbitrary graph corruption.
\subsection{Memory-Aware Model Prediction}
\label{subsec:dualmodel}
While temporal-aware augmentation provides diverse input views, the absence of ground-truth labels poses a fundamental challenge to online adaptation. Without supervision, the model risks overfitting to noisy or misaligned predictions. To overcome this, we employ a dual-model prediction scheme consisting of: (a) a well-trained DGNN model that remains fixed during adaptation and serves as a stable source of pseudo-labels, and (b) an adaptive model that is continuously updated through consistency-based objectives to track the evolving graph distribution.
This design enables the adaptive model to benefit from the stable guidance of the well-trained DGNN while gradually aligning with test-time dynamics via online updates. The memory bank also helps reduce confirmation bias during online adaptation. If the adaptive model relies only on its current prediction, an incorrect high-confidence prediction may be immediately reinforced by the consistency objective, leading to error accumulation over time. By aggregating predictions from recent timestamps, the memory bank provides a temporally smoothed target that is less sensitive to abrupt noise at a single snapshot. 
At the same time, the exponential weighting scheme prevents outdated predictions from dominating the pseudo-label, allowing the model to remain responsive to newly emerging graph patterns.

\noindent\textbf{Drift-Aware Exponential Moving Average.} Exponential moving average (EMA) is a widely used technique for stabilizing model updates by maintaining a smoothed version of model parameters:
\(\boldsymbol{\theta} \gets \alpha \cdot \boldsymbol{\theta} + (1-\alpha) \cdot \boldsymbol{\theta}\), where $\alpha$ is the momentum coefficient that controls how much past parameters influence the update. While EMA based models are highly effective in static domains, they become vulnerable in dynamic graphs. In particular, EMA `misremembers' outdated patterns or generates unstable pseudo-labels when the graph structure changes drastically — e.g., when new edges emerge or previous neighbors disappear. Instead of using a fixed momentum, we modulate the EMA coefficient by the
structural drift between two successive graph snapshots:
\begin{equation}
\boldsymbol{\theta}_{t} \;\gets\;
\alpha_t\,\boldsymbol{\theta}_{t-1} + \bigl(1-\alpha_t\bigr)\,\boldsymbol{\theta}_t,
\label{eq:ema_update}
\end{equation}
where the time-dependent momentum $\alpha_t$ is defined as
\begin{equation}
\alpha_t \;=\; \alpha_{\text{base}}\,
\exp(-\gamma\,d_t),\qquad
\alpha_{\text{base}}\in(0,1),\;\gamma>0.
\label{eq:adaptive_alpha}
\end{equation}

\noindent
The drift term $d_t$ quantifies the extent of graph changes between time steps $t-1$ and $t$, and can be defined using either of the following two interchangeable formulations.

\noindent(a) \textit{Edge-set Jaccard distance.}
\begin{equation}
    d_t \;=\;
    \frac{\lvert\,\mathcal{E}_t \,\triangle\, \mathcal{E}_{t-1}\rvert}
         {\lvert\,\mathcal{E}_t \,\cup\,       \mathcal{E}_{t-1}\rvert},
    \label{eq:edge_jaccard}
\end{equation}
where $\triangle$ denotes symmetric difference.

\noindent(b) \textit{Adjacency-matrix Frobenius distance.}
\begin{equation}
    d_t \;=\;
    \frac{1}{|V|^{2}}\,
    \bigl\|\mathbf{A}_t - \mathbf{A}_{t-1}\bigr\|_{F}^{2},
    \label{eq:adj_fro}
    \end{equation}
with $\mathbf{A}_t$ the binary adjacency matrix at time $t$.

A larger $d_t$ (i.e., more drastic structural change) yields a smaller $\alpha_t$, which causes the adaptive model to rely less on the fixed well-trained DGNN and update more aggressively—mitigating the risk of `misremembering' obsolete structural patterns. Conversely, when the graph is relatively stable, $\alpha_t$ stays close to the base momentum $\alpha_{\text{base}}$, allowing the adaptive model to retain long-term knowledge from the well-trained DGNN.

\noindent\textbf{Memory-Bank Pseudo Labeling.} To reduce prediction variance and enhance stability during online adaptation, we maintain a fixed-length \emph{memory bank} that stores the most recent $K{+}1$ predictions from the evolving model:
\(
\{ \mathbf{y}_{t}, \mathbf{y}_{t-1}, \dots, \mathbf{y}_{t-K} \}.
\)
Here, $\mathbf{y}_{t-k}$ represents the predicted logits (or probabilities) produced by the online adaptive model at time $t{-}k$, which was initialized from the well-trained DGNN. By aggregating historical predictions, the memory bank provides temporal context that smooths label estimation and mitigates the impact of noisy fluctuations in any single time step.

We construct the memory-enhanced pseudo label as an exponentially weighted average of past predictions:
\begin{equation}
\hat{\mathbf{y}}_t = \sum_{k=0}^{K} w_{t-k}\,\mathbf{y}_{t-k},\qquad
w_k = \frac{e^{-\lambda k}}{\sum_{j=0}^{K} e^{-\lambda j}},
\label{eq:memory_pseudo}
\end{equation}
where $\lambda > 0$ controls the decay rate, giving more weight to recent predictions while down-weighting older ones.

To further stabilize the pseudo-supervision signal, we blend the memory-aggregated prediction $\hat{y}$ with the recent output $\mathbf{y}_t$ of the online model using a convex interpolation:
\begin{equation}
\tilde{\mathbf{y}}_t = \omega \cdot \hat{\mathbf{y}} + (1 - \omega) \cdot \mathbf{y}_t,
\label{eq:ref_prob}
\end{equation}
where $\omega \in [0,1]$ is a fixed mixing coefficient. This formulation ensures that the pseudo label reflects long-term historical trends (via memory). The final pseudo label $\tilde{\mathbf{y}}_t$ is then used as the reference target to guide the online model via consistency-based adaptation objectives.
\subsection{Consistency-Guided Online Adaptation}
\label{subsec:consistency}

Most existing test-time adaptation methods assume i.i.d.\ samples or stationary feature shifts, which fail in dynamic graphs where both topology and features evolve continuously. To address this, our consistency-guided online adaptation explicitly aligns predictions across time, mitigating structural dependency drift and ensuring stable adaptation under gradual temporal changes. To adapt to this evolving environment,~\method~applies a consistency-guided online adaptation  strategy that enforces alignment between current predictions and temporally smoothed pseudo labels across consecutive time steps.

\noindent\textbf{Prediction Consistency Loss.} We enforce alignment between the adaptive model’s prediction and the memory bank-based pseudo label at the current time step $t$ using Kullback–Leibler (KL) divergence:
\begin{equation}
\mathcal{L}_{\text{con}} = -\sum_{i} \mathbf{y}_{t}(i)\cdot \log \mathbf{y}_{t}'(i),
\label{eq:consistency_loss}
\end{equation}
where $\mathbf{y}_t$ is the pseudo-label derived from memory-based interpolation (Eq.~\eqref{eq:memory_pseudo}), $\mathbf{y}_t'$ is the current of the adaptive model.

\noindent\textbf{Temporal Smoothness Loss.} To regularize prediction stability over time, we introduce a temporal consistency loss that penalizes deviations between the adaptive model’s current output and its previous-step output:

\begin{equation}
\mathcal{L}_{\text{temp}} = \big\| \mathbf{y}_{t}' - \mathbf{y}_{t-1}' \big\|^2.
\label{eq:temporal_loss}
\end{equation}
This encourages~\method~to produce temporally smooth predictions, unless significant distributional drift is observed, and The total loss at test time combines both terms:
\begin{equation}
\mathcal{L} = \mathcal{L}_{\text{con}} + \lambda \cdot \mathcal{L}_{\text{temp}},
\label{eq:final_loss}
\end{equation}
where $\lambda$ is a balancing coefficient. This objective allows~\method~to adapt online while preserving short-term semantic consistency and long-term temporal stability both critical under structural drift.
\subsection{Overall Optimization Procedure}
Algorithm~\ref{alg:dgotta} summarizes the overall online adaptation procedure of DGOTTA. At each test timestamp, the incoming graph snapshot is first augmented with temporal-aware feature and edge perturbations. The adaptive model then generates predictions, which are combined with memory-based pseudo labels to compute the consistency and temporal smoothness objectives. After updating the model parameters, DGOTTA further applies a drift-aware EMA update based on the structural discrepancy between consecutive graph snapshots, and stores the latest prediction in the memory bank for future adaptation.
\begin{algorithm}[h]
\caption{DGOTTA: Temporal Memory-Aware Online Test-Time Adaptation}
\label{alg:dgotta}
\begin{algorithmic}[1]
\STATE \textbf{Input:} Pretrained DGNN $f_{\boldsymbol{\theta}^*}$; Graph stream $\{G_t=(V_t, E_t, \mathbf{x}_t)\}_{t=1}^T$;
Hyperparameters: $K$, $\omega$, $\lambda_{\text{temp}}$, $\alpha_{\text{base}}$, $\gamma$, $W$, $p_{\text{drop}}$
\STATE \textbf{Output:} Adapted model $f_{\boldsymbol{\theta}}$
\STATE Initialize $\boldsymbol{\theta} \leftarrow \boldsymbol{\theta}^*$; MemoryBank $\leftarrow \emptyset$
\FOR{$t = 1$ to $T$}
    \STATE $\tilde{\mathbf{x}}_t \leftarrow$ TemporalFeatureAugment($\mathbf{x}_t$) \hfill // Eq.(\ref{eq:timeaware_feat})
    \STATE $\tilde{E}_t \leftarrow$ TemporalEdgePerturb($E_t, W, p_{\text{drop}}$) \hfill // Eq.(\ref{eq:timewindow_edge})
    \STATE $\mathbf{y}_{\text{pred}} \leftarrow f_{\boldsymbol{\theta}}(\tilde{X}_t, \tilde{E}_t)$
    \STATE $\mathbf{y}_{\text{memory}} \leftarrow$ MemoryWeightedAvg(MemoryBank, $\lambda$) \hfill // Eq.(\ref{eq:memory_pseudo})
    \STATE $\tilde{\mathbf{y}} \leftarrow \omega \cdot \mathbf{y}_{\text{memory}} + (1 - \omega) \cdot \mathbf{y}_{\text{pred}}$ \hfill // Eq.(\ref{eq:ref_prob})
    \STATE Compute $L_{\text{con}}, L_{\text{temp}}, L$ \hfill // Eqs. (\ref{eq:consistency_loss},~\ref{eq:temporal_loss},~\ref{eq:final_loss})
    \STATE Update model: $\boldsymbol{\theta} \leftarrow \boldsymbol{\theta} - \eta \nabla_{\boldsymbol{\theta}} L$
    \STATE Compute graph drift $d_t$ from $E_t$, $E_{t-1}$ \hfill // Eq.(\ref{eq:edge_jaccard}) or (\ref{eq:adj_fro})
    \STATE $\alpha_t \leftarrow \alpha_{\text{base}} \cdot \exp(-\gamma \cdot d_t)$ \hfill // Eq.(\ref{eq:adaptive_alpha})
    \STATE $\boldsymbol{\theta} \leftarrow \alpha_t \cdot \boldsymbol{\theta} + (1 - \alpha_t) \cdot \boldsymbol{\theta}^*$ \hfill // Eq.(\ref{eq:ema_update})
    \STATE Update MemoryBank with $\mathbf{y}_{\text{pred}}$
\ENDFOR
\end{algorithmic}
\end{algorithm}
\section{Experiment}
\subsection{Experiment Setting}
\noindent\textbf{Dataset.} 
We conduct experiments on three real-world dynamic graph datasets: Reddit, Wikipedia (Wiki), and MOOC, which are widely used in temporal graph learning benchmarks~\cite{kumar2019predicting}. 
\begin{itemize}
    \item \textbf{Wikipedia}: consists of edits on Wikipedia pages over one month. Editors and Wiki pages are modelled as nodes, and the timestamped posting requests are edges.
    \item \textbf{Reddit}: models subreddits’ posted spanning one month, where the nodes are users or posts and the edges are the timestamped posting requests.
    \item \textbf{MOOC}: is a student interaction network formed from online course content units such as problem sets and videos. Each edge is a student accessing a content unit and has 4 features.
\end{itemize}
The details of dataset statistics are in Table~\ref{tab:dataset}.
\begin{table*}[h]
    \caption{Statistical details of the experimental datasets.}
    \centering
    \begin{tabular}{c|ccccccc}
        \toprule
         Dataset&Domain& \# Nodes& Total Edges & Unique Edges & Unique Steps & Time Granularity & Duration  \\
         \midrule
         Wikipedia& Social & 9,227 & 157,474& 18,257& 152,757 & Unix timestamp & 1 month\\
         Reddit& Social & 10,984 & 672,447 & 78,516 & 669,065 & Unix timestamp & 1 month\\
         MOOC& Interaction & 7,144 & 411,749 & 178,443 & 345,600 & Unix timestamp & 17 months\\
         \bottomrule
    \end{tabular}
    \label{tab:dataset}
\end{table*}

For each dataset, we adopt a 10\%/10\%/80\% chronological split into training, validation, and test sets. This temporal partitioning ensures that the model is trained on early graph snapshots and evaluated on future unseen dynamics, which aligns with the real-world scenario of online deployment. We choose this split to simulate the practical setting of test-time adaptation, where the model is trained on a small historical portion and must generalize to a long sequence of incoming test data with potential distribution shift. The large test portion (80\%) allows us to observe long-horizon adaptation performance and evaluate robustness under temporal drift.

\noindent\textbf{Baselines.} 
Since our method is the first online test-time adaptation approach on dynamic graph, there is no available baseline for direct comparison,  We compare our method against two categories of test-time adaptation (TTA) baselines: 
(1) a generic TTA approach, TENT~\cite{wang2020tent}, and 
(2) graph-specific TTA methods, including GTrans~\cite{jin2022empowering}, SOGA~\cite{mao2024source}, and MATCHA~\cite{bao2025matcha}. We use this baselines with another feature time to do dynamic graph test-time adaptation at each timestamp to compare fairly.
Moreover, we include a self-supervised dynamic graph learner, DDGCL~\cite{10.1145/3459637.3482389}, and a topological-aware TTA method, IDOL~\cite{10.1145/3637528.3671777}, for comprehensive comparison. To ensure a fair evaluation, we restrict our comparison to TTA settings, where each method adapts a pretrained model to a target dynamic graph during adaptation. 
We consider four DGNN as four representative backbones, i.e., GraphMixer~\cite{cong2023we} (GMixer. for short), DyGFormer~\cite{yu2023towards} (DyGF. for short), TGN~\cite{rossi2020temporal}, and TGAT~\cite{xu2020inductive} for the node classification task. 
All models are pretrained on the training dynamic graph using Empirical Risk Minimization (ERM), and test-time adaptation is applied to the evolving target graph. We evaluate model performance based on the Area Under the ROC Curve (AUC).

\noindent\textbf{Implementation and fairness protocol.}
For a fair comparison, all methods are initialized from the same ERM-pretrained DGNN backbone and evaluated on the same chronological test stream. 
No method is allowed to access test labels or future graph snapshots during adaptation. 
For static graph TTA baselines, we adapt them to the dynamic setting by applying their unsupervised adaptation objectives independently at each incoming timestamp. 
For graph transformation-based methods, the transformation is performed on the current test snapshot before prediction, while model-update-based methods update parameters online after observing the current unlabeled snapshot. All hyperparameters are selected based on the validation split, and the same backbone architecture and evaluation metric are used across all methods.
For \method, we use memory size $K=200$, interpolation factor $\omega=0.1$, temporal consistency weight $\lambda_{\text{temp}}=0.1$, EMA base momentum $\alpha_{\text{base}}=0.1$, and drift-aware decay factor $\gamma=0.1$. 
These values are selected according to validation performance and further analyzed in the parameter sensitivity study.
\begin{table*}[h!]
\centering
\caption{Test AUC on three benchmark datasets under different backbones. `OOM' indicates out-of-memory failure.}
\small
\setlength{\tabcolsep}{3pt}
\begin{tabular}{l|
    P{1cm}P{1cm}P{1cm}P{1cm}|
    P{1cm}P{1cm}P{1cm}P{1cm}|
    P{1cm}P{1cm}P{1cm}P{1cm}}
\toprule
\textbf{Methods} & \multicolumn{4}{c|}{\textbf{Wikipedia}} & \multicolumn{4}{c|}{\textbf{Reddit}} & \multicolumn{4}{c}{\textbf{MOOC}} \\\addlinespace[2pt]
 & GMixer & DyGF & TGN & TGAT & GMixer & DyGF & TGN & TGAT & GMixer & DyGF & TGN & TGAT \\
\midrule
ERM              & 0.5018 & 0.7311 & 0.5855 & 0.5397 & 0.5037 & 0.5254 & 0.5272 & 0.5566 & 0.5064 & 0.5224 & 0.5666 & 0.5824 \\
DDGCL\scriptsize{~\cite{10.1145/3459637.3482389}}   & 0.4646 & 0.4168 & 0.5181 & 0.4677 & 0.4833 & 0.5043 & 0.5048 & 0.4913 & 0.5708 & 0.5250 & 0.5354 & 0.5166 \\
IDOL~\scriptsize{\cite{10.1145/3637528.3671777}}    & 0.5335 & 0.5016 & 0.4379 & 0.5306 & 0.5112 & 0.5083 & 0.5420 & 0.5133 & 0.5738 & 0.4930 & 0.4932 & 0.4973 \\
TENT~\scriptsize{\cite{wang2020tent}}             & 0.5140 & 0.5055 & 0.5175 & 0.4999 & 0.5014 & 0.4999 & 0.5002 & 0.5011 & 0.4999 & 0.5002 & 0.5032 & 0.4999 \\
GTrans~\scriptsize{\cite{jin2022empowering}}           & 0.5175 & 0.7731 & 0.5005 & 0.4986 & OOM    & 0.4560 & OOM    & OOM    & OOM    & 0.4884 & 0.5140 & OOM    \\
SOGA~\scriptsize{\cite{mao2024source}}             & 0.5233 & 0.7736 & 0.6137 & 0.4788 & 0.5114 & 0.4684 & 0.5090 & 0.4952 & 0.5712 & 0.5177 & 0.5032 & 0.5222 \\
MATCHA~\scriptsize{\cite{bao2025matcha}}       & 0.5259 & 0.7788 & 0.6137 & 0.5509 & 0.4785 & 0.5139 & 0.5481 & 0.5984 & 0.4931 & 0.4995 & 0.5130 & 0.4986 \\
\midrule
\textbf{\method (ours)} & \textbf{0.5827} & \textbf{0.7923} & \textbf{0.6433} & \textbf{0.5614} & \textbf{0.6036} & \textbf{0.5613} & \textbf{0.5836} & \textbf{0.6011} & \textbf{0.6233} & \textbf{0.5428} & \textbf{0.6133} & \textbf{0.6212} \\
\bottomrule
\end{tabular}
\label{tab:main_results}
\end{table*}
\subsection{Main Results}
Table~\ref{tab:main_results} reports the AUC scores of different test-time adaptation methods across three dynamic graph benchmarks and four DGNN architectures. Overall, our proposed method \method~ consistently achieves the best performance across all datasets and backbones, highlighting its effectiveness in handling both temporal and structural distribution shifts. 

On the Wikipedia dataset, \method~achieves the highest AUC on all backbones, with notable improvements of +8.1\% on GraphMixer, +6.1\% on TGN, and +2.2\% on TGAT compared to the best-performing baseline. On Reddit, while some baselines such as GTrans fail due to out-of-memory (OOM) issues on certain backbones, \method~remains robust and achieves state-of-the-art results (e.g., 0.6036 on GraphMixer and 0.6011 on TGAT). Similarly, on MOOC, \method~achieves consistent gains across all architectures, demonstrating strong generalization even in the presence of sparse and noisy temporal interactions.

Traditional methods like ERM and TENT show limited adaptability in dynamic graphs, reflecting their inability to update representations under evolving graph distributions. Graph-specific TTA methods such as GTrans and SOGA improve over ERM in some cases but suffer from poor stability across datasets, suggesting that they fail to fully capture temporal dynamics. MATCHA, while benefiting from consistency-based self-training, is less effective under severe temporal drifts, especially on Reddit compared to Wikipedia.

These results demonstrate that \method~is not only effective but also model-agnostic: it provides consistent performance gains across diverse DGNN backbones without retraining or hyperparameter tuning on specific datasets. This robustness, combined with its scalability, underlines the practical value of \method~for real-world dynamic graph applications where temporal distribution shifts are inevitable.

\noindent\textbf{Backbone-wise observations.}
We further observe that the effectiveness of \method is consistent across different types of DGNN architectures. 
For attention-based backbones such as TGAT and DyGFormer, \method improves adaptation by providing temporally smoothed pseudo-labels and reducing prediction inconsistency across consecutive snapshots. 
For memory-based or mixer-style models such as TGN and GraphMixer, \method further complements their temporal representation learning by explicitly modeling test-time structural drift and pseudo-label stability. 
This suggests that \method does not rely on a specific architectural design of the backbone. 
Instead, it acts as a general online adaptation layer that can be integrated with different DGNNs to improve robustness under evolving test distributions.

Another important observation is that several existing baselines perform well on certain datasets but are unstable across backbones. 
For example, some graph-specific TTA methods achieve competitive results on Wikipedia but degrade on Reddit or MOOC, indicating that adaptation strategies designed for static graph shifts may not generalize well to dynamic graph streams. 
In contrast, \method consistently achieves the best or near-best performance across all backbone-dataset combinations, demonstrating stronger robustness to both temporal and structural shifts.
\begin{figure*}[!t]
    \centering
    \begin{subfigure}[b]{0.6\textwidth}
        \centering
        \includegraphics[width=\textwidth]{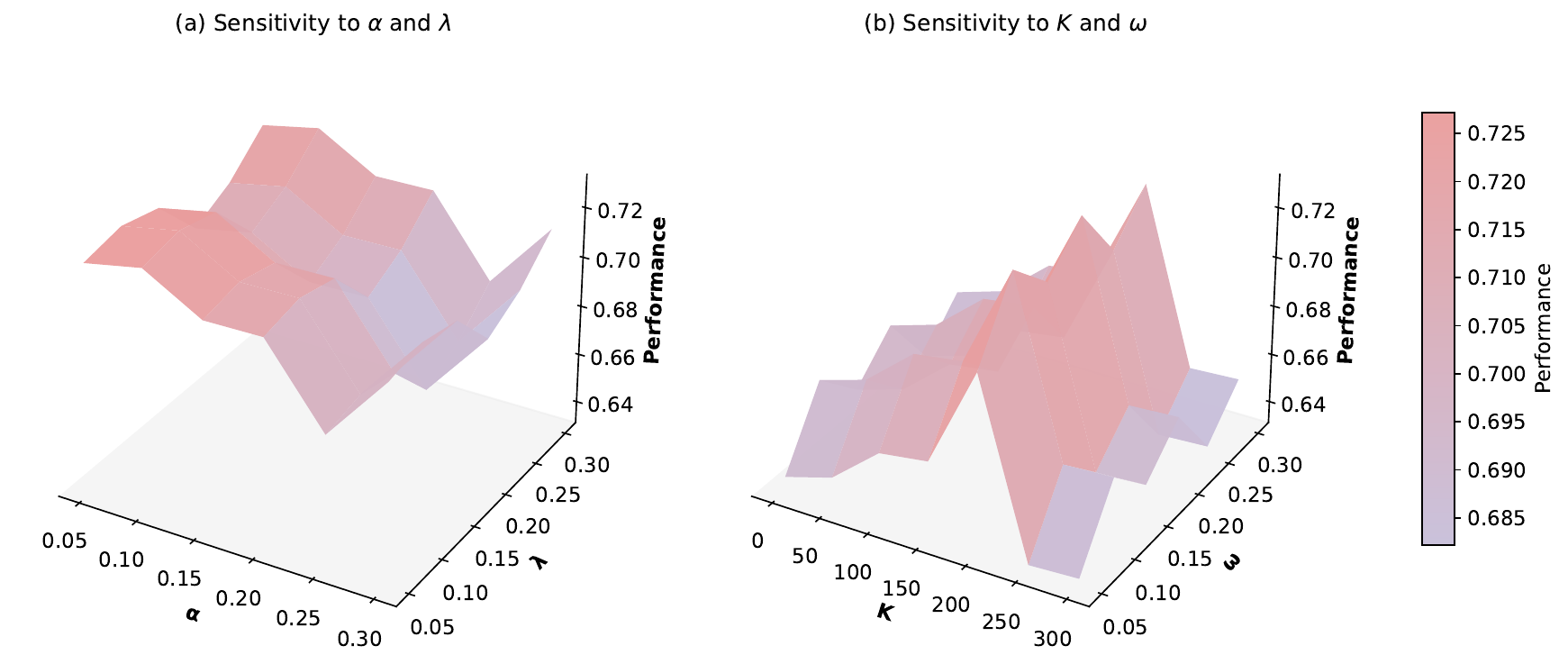}
        \caption{Parameter sensitivity surfaces of DyGFormer.}
        \label{fig:param_sensitivity}
    \end{subfigure}
    \hfill
    \begin{subfigure}[b]{0.39\textwidth}
        \centering
        \includegraphics[width=\textwidth]{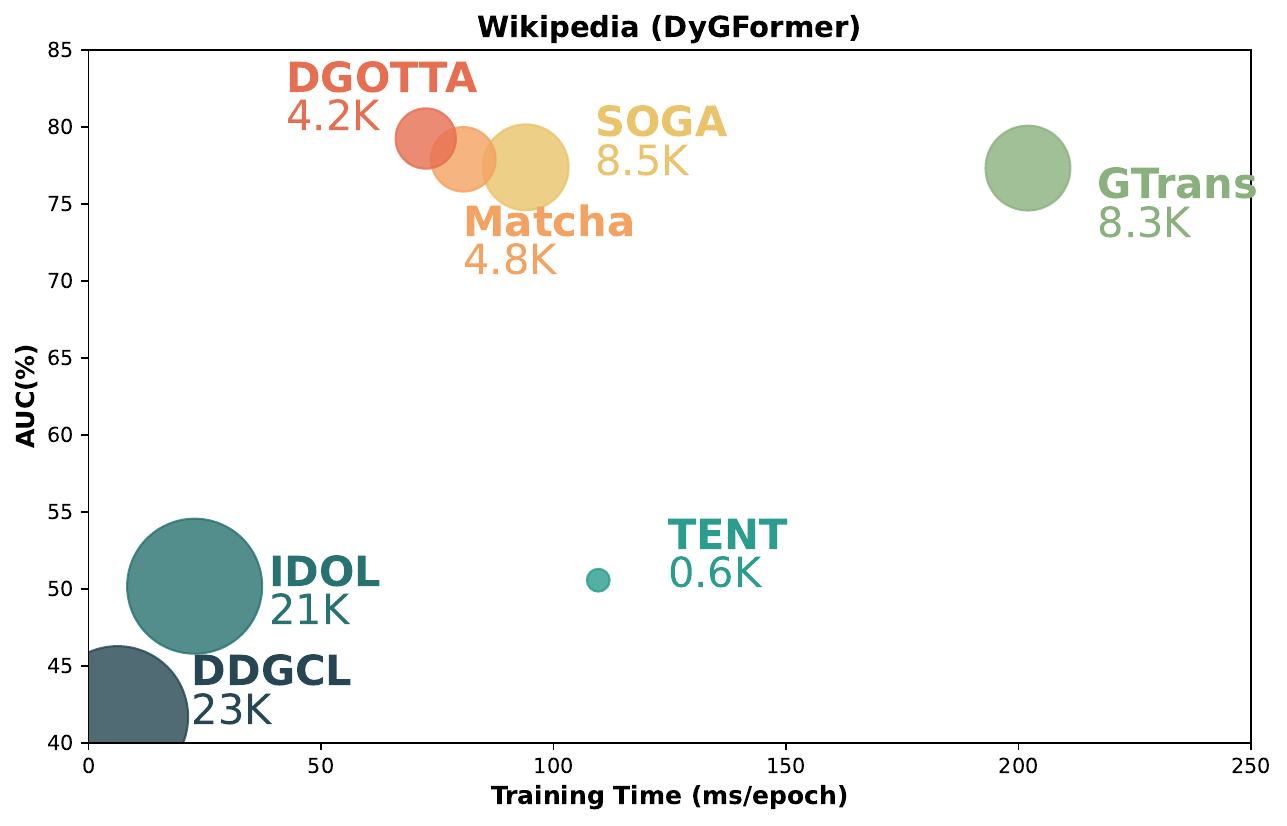}
        \caption{Runtime–AUC trade-off of DGOTTA.}
        \label{fig:efficiency_scatter}
    \end{subfigure}
    \caption{Parameter sensitivity and efficiency analysis.}
    \label{fig:parameter_and_efficiency}
\end{figure*}
\subsection{Ablation Study}

\begin{table*}[t]
\centering
\caption{Ablation study under different backbones.~\method-base is our proposed method without temporal consistency loss (TC Loss), memory-bank pseudo label (MBPL), temporal-aware augmentation (TAA), and drifter-aware EMA (DAEMA).}
\small
\setlength{\tabcolsep}{5pt}
\begin{tabular}{l|cccc|cccc|cccc}
\toprule
\textbf{Variants} & \multicolumn{4}{c|}{\textbf{Wikipedia}} 
& \multicolumn{4}{c|}{\textbf{Reddit}} 
& \multicolumn{4}{c}{\textbf{MOOC}} \\\addlinespace[2pt]
 & GMixer & DyGF & TGN & TGAT 
 & GMixer & DyGF & TGN & TGAT 
 & GMixer & DyGF & TGN & TGAT \\
\midrule
\method-base         & 0.5009 & 0.6008 & 0.5009 & 0.5197 & 0.5012 & 0.5302 & 0.5212 & 0.5499 & 0.5005 & 0.5008 & 0.5003 & 0.5233 \\
+TC Loss             & 0.5043 & 0.5976 & 0.5239 & 0.5105 & 0.5025 & 0.5289 & 0.5267 & 0.5149 & 0.5018 & 0.5185 & 0.5021 & 0.5338 \\
+MBPL                & 0.4943 & 0.7658 & 0.5232 & 0.5097 & 0.5031 & 0.5287 & 0.5267 & 0.5350 & 0.5030 & 0.5069 & 0.5099 & 0.5384 \\
+TAA                 & 0.5442 & 0.7471 & 0.5950 & 0.5362 & 0.5520 & 0.5308 & 0.5565 & 0.5535 & 0.5472 & 0.5077 & 0.5131 & 0.5158 \\
+DAEMA               & 0.5708 & 0.6998 & 0.6186 & 0.5507 & 0.5730 & 0.5339 & 0.5729 & 0.5518 & 0.5651 & 0.5081 & 0.5083 & 0.5445 \\\midrule
\textbf{\method (ours)}     & \textbf{0.5827} & \textbf{0.7923} & \textbf{0.6433} & \textbf{0.5614} 
& \textbf{0.6036} & \textbf{0.5613} & \textbf{0.5836} & \textbf{0.6011} 
& \textbf{0.6233} & \textbf{0.5428} & \textbf{0.6133} & \textbf{0.6212} \\
\bottomrule
\end{tabular}
\label{tab:ablation}
\end{table*}
We conduct an ablation study to assess the contribution of individual components in DGOTTA, with results summarized in Table~\ref{tab:ablation}.
Starting from DGOTTA-base, which applies a pretrained DGNN directly to the test graph stream without adaptation, we observe clear performance degradation under temporal and structural distribution shifts.

Adding the temporal consistency (TC) loss provides modest but consistent improvements, indicating that enforcing prediction smoothness helps reduce noisy updates in dynamic settings.
Incorporating memory-bank pseudo labeling (MBPL) leads to substantial gains by stabilizing pseudo labels through historical prediction aggregation.
Temporal-aware augmentation (TAA) improves performance, demonstrating the importance of modeling temporal variations in node features and interaction structures.
The drift-aware EMA (DAEMA) module brings additional benefits by adapting the update rate to structural changes, thereby mitigating catastrophic forgetting.

Combining all components yields the best performance across datasets and backbones, confirming that the proposed modules are complementary and jointly enable robust online test-time adaptation on dynamic graphs.
\subsection{Analysis of Distribution Shift Mitigation.}
We further analyze the ablation results from the perspective of distribution shift mitigation in dynamic graphs. Each component in DGOTTA is designed to address a specific challenge induced by temporal distribution shifts.

Specifically, temporal-aware augmentation (TAA) targets semantic and temporal feature shifts by perturbing node representations according to interaction recency, thereby reducing the mismatch between training-time and test-time feature distributions. The consistent performance gains observed after adding TAA indicate that explicitly modeling temporal feature drift is crucial under evolving graph streams.

The memory-based pseudo-labeling module is designed to alleviate prediction instability caused by non-stationary graph dynamics. As shown in the ablation results, incorporating memory aggregation leads to more stable improvements, particularly on datasets with pronounced temporal variability, suggesting its effectiveness in smoothing noisy predictions under distribution shifts.

Finally, the drift-aware EMA mechanism explicitly addresses catastrophic forgetting in online adaptation. Unlike conventional EMA schemes that assume stationary update dynamics, the proposed drift-aware EMA adjusts the update rate according to structural changes between consecutive snapshots. The additional gains brought by this component demonstrate its importance in preserving useful historical knowledge while adapting to evolving graph structures.

Another observation is that the components are not isolated from each other. 
Temporal-aware augmentation increases the diversity of test-time views, but this may also introduce noisy predictions if used without stable pseudo-supervision. 
The memory bank alleviates this issue by aggregating recent predictions into smoother targets. 
Similarly, drift-aware EMA controls how strongly the model should retain historical knowledge under different levels of structural change. 
Therefore, the full model benefits from the interaction among augmentation, memory smoothing, and drift-aware updating, which explains why combining all modules achieves the strongest performance.

Overall, the ablation results confirm that the performance improvements of DGOTTA stem from its ability to mitigate multiple types of distribution shifts in dynamic graphs, rather than from generic regularization effects.
\subsection{Parameter Analysis}
We conduct a parameter sensitivity analysis to investigate several key hyperparameters, including the memory bank size $K$, reference-adaptation interpolation factor $\omega$, temporal consistency weight $\lambda$, and EMA update ratio $\alpha$. Figure~\ref{fig:param_sensitivity} shows the AUC performance on the Wikipedia dataset under different parameter settings.

\noindent\textbf{Memory Size $K$.}
As shown in the first plot, increasing the memory bank size $K$ leads to performance gains for DyGFormer up to a peak at $K=200$, while GraphMixer achieves the best performance around $K=200$ but drops sharply after that. This indicates that a moderate memory size helps stabilize pseudo-labels, while excessive memory may introduce outdated or noisy predictions.

\noindent\textbf{Interpolation Factor $\omega$.}
 A smaller $\omega$ (i.e., giving more weight to memory bank predictions) leads to improved results, with optimal performance around $\omega=0.1$. This confirms the effectiveness of aggregating historical knowledge.

\noindent\textbf{Temporal Consistency Weight $\lambda$.}
From the third plot, we find that setting $\lambda=0.1$ yields the best performance. This confirms that enforcing temporal smoothness helps the adaptive model generalize better across evolving graph structures, but excessive regularization may hinder flexibility.

\noindent\textbf{EMA Update Ratio $\alpha$.}
In the fourth plot, performance peaks around $\alpha=0.1$, which balances stability and adaptability. Too small $\alpha$ slows adaptation, while too large $\alpha$ makes well-trained model overfit noisy test samples.

Overall, the consistent trend across different architectures verifies the robustness of our hyperparameter design.
\subsection{Efficiency Analysis}



To evaluate the efficiency of different test-time adaptation methods, we report the average inference time per timestamp together with the corresponding AUC scores on the Wikipedia dataset using DyGFormer as the backbone.
The results are shown in Figure~\ref{fig:efficiency_scatter}, where each method is represented by a point reflecting its accuracy--efficiency trade-off.

Conventional methods such as DDGCL and IDOL achieve relatively low inference latency but exhibit limited adaptation performance, indicating a weak capability to handle dynamic structural changes.
In contrast, graph-based TTA methods, including GTrans and SOGA, improve performance at the cost of increased computational overhead.

Our proposed DGOTTA achieves the best trade-off between efficiency and performance.
Specifically, DGOTTA attains the highest AUC while maintaining a competitive inference time per timestamp, outperforming existing methods in both accuracy and runtime.
These results demonstrate that DGOTTA provides an effective and scalable solution for online adaptation in dynamic graph scenarios.
\section{Conclusion}
In this paper, we proposed~\method, a novel framework for online test-time adaptation on dynamic graphs. 
\method~is designed to handle both temporal and structural distribution shifts that commonly occur in real-world dynamic environments. By combining time-aware augmentation, a memory-aware model prediction, and consistency-guided online adaptation, \method~enables robust and efficient adaptation. Extensive experiments on real-world datasets and dynamic GNN backbones demonstrate~\method~consistently outperforms existing test-time adaptation and self-supervised baselines, achieving state-of-the-art performance with minimal overhead. We believe~\method~provides a promising direction for future research on adaptive learning in dynamic graph scenarios, particularly under realistic online inference settings.
\section*{Acknowledgments}
    This research is supported by the U.S. National Science Foundation and Australia CSIRO joint project: NSF-CSIRO: Towards Interpretable and Responsible Graph Modeling for Dynamic Systems (IIS-2302786). This work has also been supported by the Australian Research Council (ARC) under grants FT210100097 and DP240101547. Shirui Pan, Xin Zheng, and Ming Jin acknowledge the support of the 2025 NVIDIA Academic Grants Program for the project “Multimodal Knowledge-Aware Learning with LLMs”.

\bibliography{reference}
\bibliographystyle{IEEEtran}

\end{document}